%% file: Template.tex
\documentclass{article}
\usepackage{spconf,amsmath,graphicx,hyperref}
\usepackage{cite}
\usepackage{amsmath,amssymb,amsfonts}

\usepackage{graphicx}
\usepackage{textcomp}
\usepackage{xcolor}
\usepackage{tabularx}
\usepackage[numbers, sort&compress]{natbib}
\usepackage{caption}
\usepackage{booktabs}
\usepackage{multirow}
\usepackage{hyperref}
\usepackage{tipa}
\usepackage{float}

\usepackage[ruled,vlined,linesnumbered]{algorithm2e}
\SetKw{Continue}{continue}

\title{K-Function: Joint Pronunciation Transcription and Feedback for Evaluating Kids Language Function}
\name{
    \begin{tabular}{@{}c@{}}
    Shuhe Li$^{1}$\textsuperscript{*} \quad 
    Chenxu Guo$^{1}$\textsuperscript{*} \quad 
    Jiachen Lian$^{2}$\textsuperscript{†} \quad 
    Cheol Jun Cho$^{2}$ \quad 
    Wenshuo Zhao$^{1}$ \quad 
    Xiner Xu$^{1}$ \\
    \textit{Ruiyu Jin}$^{1}$ \quad 
    \textit{Xiaoyu Shi}$^{3}$ \quad
    \textit{Xuanru Zhou}$^{1}$ \quad
    \textit{Dingkun Zhou}$^{4}$ \quad 
    \textit{Sam Wang}$^{2}$ \quad 
    \textit{Grace Wang}$^{2}$ \\
    \textit{Jingze Yang}$^{1}$ \quad 
    \textit{Jingyi Xu}$^{1}$ \quad 
    \textit{Ruohan Bao}$^{1}$ \quad
    \textit{Xingrui Chen}$^{5}$ \quad
    \textit{Elise Brenner}$^{6}$ \quad 
    \textit{Brandon In}$^{6}$ \\
    \textit{Francesca Pei}$^{6}$ \quad
    \textit{Maria Luisa Gorno-Tempini}$^{6}$ \quad 
    \textit{Gopala Anumanchipalli}$^{2}$
    \end{tabular}
    \thanks{* Equal contribution. † Contact: \texttt{jiachenlian@berkeley.edu}}
}

\address{$^{1}$Zhejiang University \quad $^{2}$UC Berkeley \quad $^{3}$Duke University \quad $^{4}$SCUT \quad $^{5}$TVT \quad $^{6}$UCSF}
\begin{document}
\ninept
\maketitle
\begin{abstract}
Evaluating young children's language is challenging for automatic speech recognizers due to high-pitched voices, prolonged sounds, and limited data. We introduce K-Function, a framework that combines accurate sub-word transcription with objective, Large Language Model (LLM)-driven scoring. Its core, Kids-Weighted Finite State Transducer (K-WFST), merges an acoustic phoneme encoder with a phoneme-similarity model to capture child-specific speech errors while remaining fully interpretable. K-WFST achieves a 1.39\% phoneme error rate on MyST and 8.61\% on Multitudes—an absolute improvement of 10.47\% and 7.06\% over a greedy-search decoder. These high-quality transcripts are used by an LLM to grade verbal skills, developmental milestones, reading, and comprehension, with results that align closely with human evaluators. Our findings show that precise phoneme recognition is essential for creating an effective assessment framework, enabling scalable language screening for children. Our K-Function framework demo is available at \url{https://chenxukwok.github.io/K-function/}.
\end{abstract}
\section{Introduction}

\input{text/intro}

\section{Method}

\input{text/method}

\section{Experiments}

\input{text/experiments}

\section{Conclusion and future work}

\input{text/conclusion}

\bibliographystyle{IEEEbib}
\bibliography{strings,refs}

\end{document}

%% file: text/intro.tex
1 in 14 children in the U.S. has a speech or language disorder, such as dyslexia or Autism, that can hinder reading, vocabulary growth, and communication~\cite{nidcd2023, cde2024, crossriver2024, gorno2011classification-ppa}. These delays can affect quality of life, academic success, and future opportunities. Thus, developing a reliable and automatic framework for evaluating kids language function, covering verbal skills, developmental milestones, reading, and comprehension, is essential. It enables early detection of delays and supports educators and specialists in early identification and intervention.

The bottleneck in developing such a framework lies in the accurate transcription of kids speech, which is significantly more challenging than that of adults. This difficulty arises from several unique characteristics of child vocalizations, including higher fundamental frequency, longer phone durations, greater articulatory variability, and persistent data sparsity~\cite{lee1997analysis, tran2020analysis, neuberger2014cross, gerosa2009review, yeung2018difficulties, shankar2025selective}. 

Although recent work has produced increasingly robust child-oriented Automatic Speech Recognition (ASR) models~\cite{fan2024benchmarking, bhardwaj2022automatic, shahnawazuddin2024developing, singh2025causal, singh2021data, kathania2022formant, fan2022towards, xu24c_interspeech, kim2021conditionalvariationalautoencoderadversarial}, most developmentally significant pronunciation errors arise at the sub-word level and are consequently hidden by word-level metrics such as Word Error Rate (WER). Consider the prompt bird /\textipa{b@rd}/: a child who utters /\textipa{bed}/ (“bed”) exhibits a phoneme substitution, while /\textipa{pb@rd}/ (“p-bird”) contains an intrusive plosive. State-of-the-art end-to-end ASR systems often output the canonical word bird for both productions, masking the underlying articulatory deviation. A dedicated \textit{sub-word recognizer} that reliably recovers phoneme sequences is therefore indispensable for fine-grained kids language function evaluation, where \textit{phoneme recognition} is the most widely used approach.

Traditional work has primarily focused on fine-tuning children's speech data on pretrained self-supervised speech learning (SSL) models~\cite{li2024enhancingchildvocalizationclassification, li2024analysisselfsupervisedspeechmodels, Shi-JASAEL-2024, gao2024g2pu, peng2023study, 10971198, zhu2022phone-w2v2-alignment, zhao-etal-2025-mutis, mohamed2022selfreview, hsu2021hubert, Chen_2022}. However, these approaches exhibit poor generalizability to open-domain children's speech. Moreover, directly adopting adult-oriented ASR or phoneme recognition pipelines fails to address allophonic variations and disrupted phoneme-level language structures. 

Unlike purely data-driven approaches, the real breakthrough in modeling atypical speech transcription lies in the development of disfluent speech transcription pipelines that incorporate human and linguistic priors~\cite{ssdm, zwilling2025speech, ye2025seamlessalignment-neurallcs, zhang2025analysisevaluationsyntheticdata, lian22b_interspeech-gestures, lian2023articulatory-factors}. A state-of-the-art example is Dysfluent Weighted Finite State Transducer (D-WFST), which leverages WFST to model disrupted phoneme structures~\cite{guo2025dysfluentwfstframeworkzeroshot}. However, these WFST-based approaches lack robustness in detecting subtle phonetic variations, particularly substitutions and deletions, an issue especially pronounced in downstream applications involving audio with moderate to severe disfluency.

In this work, we present K-Function, a comprehensive framework for evaluating children’s language function. It consists of our newly proposed Kids-Weighted Finite State Transducer (K-WFST), a lightweight and interpretable kids phoneme transcription model—an LLM-boosted automatic language function scoring system, and a feedback component. Together, these modules form a unified assessment framework for generalized language function evaluation in children.

K-Function is evaluated on two child speech datasets MyST~\cite{ward2011my} and Multitudes~\cite{UCSFMultitudes2025}. Our framework consistently achieves superior phoneme recognition, enables more accurate language function scoring via Large Language Models (LLMs), and aligns closely with real proctor scores.

Crucially, the high-fidelity transcriptions from K-WFST provide the foundation for a complete assessment-feedback loop, powering sophisticated downstream analyses. These results collectively highlight that accurate sub-word recognition is a crucial, missing cornerstone for kids language function evaluation. Our K-Function makes the following key contributions:

\begin{figure*}[h]
\vspace{-2mm}
    \centering
     \includegraphics[width=\textwidth]{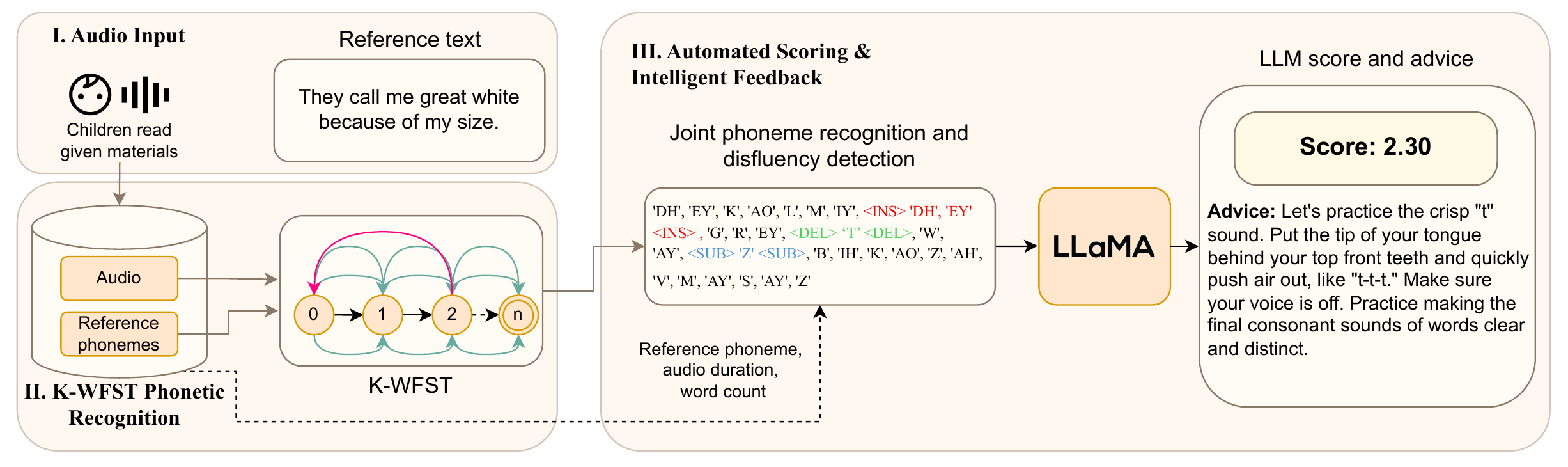}
    \caption{The 3-stage pipeline of the K-Function framework, from audio input to a comprehensive feedback report.}
    \label{fig:pipeline}
\vspace{-5mm}
\end{figure*}

\textit{(1) K-WFST:} We present a WFST decoder designed specifically for children’s speech, which achieves state-of-the-art phoneme error rates (PER) on both the MyST~\cite{ward2011my} and Multitudes~\cite{UCSFMultitudes2025} corpora.

\textit{(2) A Unified Assessment Framework:} We propose a complete system that integrates our state-of-the-art sub-word transcription (K-WFST) with an advanced LLM-based scoring module to create a robust pipeline for automated language function assessment.

\textit{(3) Align with Proctor Scores and Go Beyond:} Our framework's automated scores show high consistency with the manual scores given by expert proctors on the Multitudes corpus. Critically, our system also provides detailed, phoneme-level error analysis, offering insights that are difficult to achieve at scale through manual scoring alone.

\textit{(4) Demonstrated Downstream Utility:} We show significant value in practical applications, including achieving higher consistency in LLM-assisted language function assessment.

%% file: text/method.tex
Our proposed K-Function framework is an end-to-end pipeline designed for child speech assessment, converting raw audio into a comprehensive evaluation report. The architecture, depicted in Figure~\ref{fig:pipeline}, consists of three primary stages: audio input, K-WFST transcription, and automated scoring. 


Our pipeline consists of three main stages, each corresponding to a subsection below. First, in Datasets and Preprocessing, we describe the data sources and preparation steps needed to adapt the system to child speech. Second, in K-WFST Phonetic Recognition, we introduce our enhanced phonetic recognition framework that improves robustness to child-specific variations. Finally, in LLM-based Scoring, we detail how the recognized phoneme sequences are used with a large language model to produce automated evaluation scores aligned with expert assessments.

\subsection{Datasets and Preprocessing}
Our study utilizes two distinct datasets for model fine-tuning and evaluation: the My Science Tutor (MyST)~\cite{ward2011my} dataset for adapting our acoustic model, and the UCSF California Multitudes~\cite{UCSFMultitudes2025} corpus  for evaluating downstream performance.

\subsubsection{MyST Dataset}
For fine-tuning our base acoustic model to the unique characteristics of child speech, we use the MyST dataset, which contains conversational speech from students in third to fifth grade (8-10 years old) interacting with a virtual tutor. To align with the shorter vocalizations typically produced by younger children, we select transcribed utterances with a duration of under 20 seconds.

To generate phoneme-level labels required for end-to-end model training, we process the reference texts using the nltk~\cite{bird2009natural} toolkit to convert them into phoneme sequences. The resulting dataset is partitioned into 61.5 hours for training and 11.4 hours for testing.

\subsubsection{UCSF California Multitudes Corpus}
To evaluate our framework's practical utility in downstream assessment tasks, we use data from the UCSF California Multitudes corpus. This corpus is sourced from a digital universal screener administered to a representative sample of K-2 California public school children.

Our study specifically utilizes data from the Oral Reading Fluency (ORF) task, in which a child reads a passage aloud for two minutes. The passages read by the children are drawn from nine different reading materials: Grizzly, Banana, Quail, Raccoon, Shark, Lizard, Condor, Fox, and Sealion. A key feature of this dataset is that each ORF performance is manually scored by trained proctors, providing a ground-truth expert score for our downstream scoring evaluation. To establish a ground-truth for our transcription model's performance on this data, we performed manual annotation to obtain the reference phoneme sequences for all audio samples used from the ORF task.

\subsection{K-WFST: Phonetic Recognizer for Disorder Screening}

Weighted Finite State Transducer (WFST) is a foundational tool in speech recognition, representing a finite-state machine that maps input sequences to output sequences. A WFST consists of a set of states and weighted arcs, where each arc is a tuple defined by a start state, an end state, an input label, an output label, and a weight. A Finite State Acceptor (FSA) is a special case of a WFST where the input and output labels on each arc are identical, effectively accepting or rejecting an input sequence rather than transducing it.

A core limitation of traditional WFST-based decoders, especially in downstream applications involving child speech, is their reduced robustness in detecting subtle sub-word variations like substitutions and deletions. To address this, we introduce K-WFST, a framework that enhances the standard Dysfluent-WFST~\cite{guo2025dysfluentwfstframeworkzeroshot} by integrating a novel phoneme similarity-based substitution structure. The central innovation is to augment the WFST graph with additional, weighted paths that represent phonetically plausible substitutions. This allows the decoder to align an audio signal to a sequence that may be slightly erroneous but is phonetically similar to the reference, which is critical for accurately transcribing variable child speech.

The process for generating this augmented graph is formally detailed in Algorithm~\ref{alg:build_graph}. The algorithm's inputs are the reference phoneme ID sequence $R$, a hyperparameter $\beta$ that controls the penalty for errors, and the SimMatrix~\cite{zhou2025phonetic-error-detection}. The SimMatrix is a pre-calculated $N \times N$ matrix where $N$ is the number of phonemes in the vocabulary. We construct SimMatrix using a heuristic approach based on eight phonological features like vowel height, voicing. The similarity between $p_i$ and $p_s$ is computed as the normalized weighted sum of their matching features, yielding a score in $[0, 1]$ that captures linguistically grounded proximity. The algorithm initializes an empty set of arcs, $L$, and iterates through all start states $i$ and end states $j$ in the reference sequence, skipping cases where $i=j$ to prevent self-loops. From $\beta$, it derives weights for correct transitions ($\alpha$) and errors ($w_{\text{err}}$). It then populates $L$ with arcs for correct paths, substitutions, deletions, and repetitions, with weights determined by $\alpha$, $w_{\text{err}}$, and the similarity scores.

\vspace{0mm}
\begin{algorithm}[htbp]
\DontPrintSemicolon
\caption{Building the Augmented WFST Graph} 
\label{alg:build_graph}

\SetKwInOut{Input}{Input}
\SetKwInOut{Result}{Result}

\Input{$R$ (Phoneme ID Sequence), $\beta$, SimMatrix}
\Result{String definition of the FSA graph}

$L \leftarrow \emptyset$ \tcp*{Initialize set of arcs}
$\alpha \leftarrow 1 - 10^{-\beta}$\;
$w_{\text{err}} \leftarrow 1 - \alpha$\;

\For{$i \leftarrow 0$ \KwTo $\text{len}(R) - 1$}{
    $p_i \leftarrow R[i]$\;
    \For{$j \leftarrow 0$ \KwTo $\text{len}(R)$}{
        \If{$i = j$}{\Continue}
        
        \If{$j = i + 1$}{
            $L \leftarrow L \cup \{(i, j, p_i, p_i, \alpha)\}$ \tcp*{Correct path}
            
            \For{\text{each similar phoneme } $p_s$ \text{ to } $p_i$}{
                $w_{\text{sub}} \leftarrow w_{\text{err}} \times (1 - \text{SimMatrix}[p_i, p_s])$\;
                $L \leftarrow L \cup \{(i, j, p_s, \text{``sub"}, w_{\text{sub}})\}$ \tcp*{Substitution path}
            }
        }
        \ElseIf{$j > i$}{
            $w_{\text{dyn}} \leftarrow w_{\text{err}} \times e^{-|i-j|^2/2}$\;
            $L \leftarrow L \cup \{(i, j, 0, \text{``del"}, w_{\text{dyn}})\}$ \tcp*{Deletion path}
        }
        \ElseIf{$j < i$}{
            $w_{\text{dyn}} \leftarrow w_{\text{err}} \times e^{-|i-j|^2/2}$\;
            $L \leftarrow L \cup \{(i, j, 0, \text{``rep"}, w_{\text{dyn}})\}$ \tcp*{Repetition path}
        }
    }
}
$L \leftarrow L \cup \{(\text{len}(R), \text{len}(R)+1, -1, -1, 0)\}$ \tcp*{Final state}
\KwRet \text{FormatToString}(L)\;
\end{algorithm}
\vspace{-2mm}

Furthermore, to control the flexibility of this substitution mechanism, we employ a Task-dependent K-Selection strategy. This allows the decoder to operate in two distinct modes depending on the input speech characteristics and the base acoustic model's performance:
$$
\text{K-Selection} = \left\{
\begin{array}{ll}
    K=1 & \text{Constrained paths} \\
    K=2 & \text{Flexible paths}
\end{array}
\right.
$$
When $K=1$, the model is constrained to only consider the most similar phoneme, which is the phoneme itself. When $K=2$, the model is allowed to consider the top two most similar phonemes, adding the extra substitution paths to enhance robustness in more challenging scenarios. This task-dependent approach ensures optimal performance across different conditions without sacrificing efficiency.

\subsection{LLM-based Scoring}
To validate the practical utility of our high-fidelity transcriptions, we designed an experiment to assess if a Large Language Model (LLM) could replicate the nuanced scoring of human experts on the Multitudes Oral Reading Fluency (ORF) task. For this task, we evaluated a state-of-the-art instruction-tuned model, \texttt{meta-llama/Llama-3.1-70B-Instruct}~\cite{grattafiori2024llama}.

We employed a few-shot prompting strategy to guide the model in simulating the evaluation process of a human proctor. For each ORF sample to be scored, the LLM was provided with a comprehensive set of inputs\footnote{Prompt is available at \url{https://chenxukwok.github.io/K-function/}}: 

(1) The official Multitudes scoring guidelines, detailing the criteria for evaluation.

(2) The original reference text the child was asked to read.

(3) The detailed phoneme-level transcription generated by our K-WFST model.

(4) Four distinct, manually-scored examples to serve as in-context demonstrations of the scoring process.

The LLM synthesized these inputs to produce a single quantitative score reflecting each child’s reading performance. To account for the probabilistic nature of the model, we set the decoding temperature to 0.5 and performed the prediction five consecutive times for each sample. All five scores generated for each sample were then included in the final evaluation. We compared this entire set of predictions against the ground-truth expert scores from the Multitudes corpus to calculate the overall MAE and MSE.

%% file: text/experiments.tex
\subsection{Model Fine-tuning and Evaluation}
Our experimental process is designed to first adapt a pre-trained model for child speech and then rigorously evaluate its performance across different conditions and decoding strategies.

\subsubsection{Fine-tuning and Evaluation on MyST}

\begin{table}[h]
\centering
\caption{PER (\%) comparison on the MyST test set. The ``Base" models are the pre-trained Wav2Vec2.0, while ``Kids-FT" models are fine-tuned on MyST. Our proposed K-WFST framework demonstrates superior performance across both conditions.}
\vspace{-2mm}
\begin{tabular}{llc}
\toprule
\textbf{Model} & \textbf{Method} & \textbf{PER $\pm$ SD (\%) } \\
\midrule
\multirow{3}{*}{\textbf{Base}} 
& Greedy & 40.26$\pm$66.92 \\
& WFST (K=1) & 3.72$\pm$27.90 \\
& WFST (K=2) & 6.91$\pm$29.00 \\
\midrule
\multirow{3}{*}{\textbf{Kids-FT}}
& Greedy & 11.86$\pm$65.89 \\
& WFST (K=1) & \textbf{1.39$\pm$9.83} \\
& WFST (K=2) & 8.31$\pm$14.67 \\
\bottomrule
\end{tabular}
\label{tab:model-eval-results}
\vspace{-4mm}
\end{table}

\begin{table*}[h]
\vspace{-2mm}
\centering
\caption{Per-Material Phoneme Error Rate (PER, \%) on the UCSF Multitudes ORF Corpus. The flexible WFST (K=2) model demonstrates the strongest performance on the adapted Kids-FT model across all nine challenging reading passages.}
\vspace{-2mm}
\resizebox{0.85\textwidth}{!}{%
\begin{tabular}{lcccccc}
\toprule
\multirow{2}{*}{\textbf{Multitudes Materials}} & \multicolumn{3}{c}{\textbf{Base Model PER (\%)}} & \multicolumn{3}{c}{\textbf{Kids-FT Model PER (\%)}} \\
\cmidrule(lr){2-4} \cmidrule(lr){5-7}
& \textbf{Greedy} & \textbf{WFST (K=1)} & \textbf{WFST (K=2)} & \textbf{Greedy} & \textbf{WFST (K=1)} & \textbf{WFST (K=2)} \\
\midrule
Grizzly & 35.61 & 22.92 & 7.95 & 7.95 & 1.85 & \textbf{1.77} \\
Banana & 45.13 & 48.16 & 37.72 & 23.21 & 15.47 & \textbf{11.41} \\
Quail & 42.05 & 31.10 & 20.14 & 14.31 & 9.72 & \textbf{6.01} \\
Raccoon & 37.48 & 28.06 & 19.36 & 11.19 & 7.10 & \textbf{4.80} \\
Shark & 49.71 & 24.20 & 11.82 & 11.07 & 7.88 & \textbf{5.63} \\
Lizard & 43.64 & 24.20 & 11.82 & 11.07 & 7.88 & \textbf{5.63} \\
Condor & 43.35 & 28.06 & 19.36 & 11.19 & 7.10 & \textbf{4.80} \\
Fox & 54.35 & 48.16 & 37.72 & 23.21 & 15.47 & \textbf{11.41} \\
Sealion & 37.48 & 28.06 & 19.36 & 11.19 & 7.10 & \textbf{4.80} \\
\bottomrule
\end{tabular}%
}
\label{tab:multitudes_results}
\vspace{-6mm}
\end{table*}

First, we fine-tuned the pre-trained Phoneme-based Wav2Vec2.0 model~\cite{baevski2020wav2vec} using the 61.5-hour training partition of the MyST dataset. This process created our child-speech-adapted models, hereafter referred to as ``Kids-FT". We then conducted a comprehensive evaluation on the 11.4-hour MyST test set, which primarily consists of relatively fluent child speech. The performance of our baseline (``Base") and fine-tuned (``Kids-FT") models, measured by Phoneme Error Rate (PER), is presented in Table~\ref{tab:model-eval-results}.

The results clearly show the substantial benefit of fine-tuning, with the Kids-FT models significantly outperforming the Base models across all decoding methods. Notably, on this fluent speech dataset, the constrained setting of our decoder, \textbf{WFST (K=1)}, achieves the optimal performance with a PER of only \textbf{1.39\%}. This finding supports our hypothesis that for less variable speech, a more constrained decoding path prevents potential error propagation and yields higher accuracy.

\subsubsection{Evaluation on Multitudes}
To assess model performance in a more challenging and realistic downstream scenario, we evaluated all models on the 1.87-hour Multitudes corpus. This dataset contains a higher degree of disfluent child speech from the Oral Reading Fluency (ORF) task. The performance of each model configuration across the nine different reading passages of the ORF task is detailed in Table~\ref{tab:multitudes_results}.

As shown in the table, the fine-tuned ``Kids-FT Model" consistently outperforms the ``Base Model" across all reading passages, reaffirming the importance of adaptation to child speech. Crucially, on this more challenging disfluent dataset, the flexible \textbf{WFST (K=2)} configuration consistently yields the lowest Phoneme Error Rate (PER) for the fine-tuned model on every single passage. This result strongly supports our Task-dependent K-Selection strategy, demonstrating that allowing greater flexibility for phonetically plausible substitutions is essential for enhancing model robustness and accuracy in complex downstream scenarios.

\subsection{LLM-Assisted Scoring on the Multitudes Corpus}

To validate the practical utility of our high-fidelity transcriptions, we assessed if a Large Language Model (LLM) could replicate the nuanced scoring of human experts. Using the Llama-3.1-70B-Instruct model, we employed a few-shot prompt that included four scored examples and the official scoring guidelines. The LLM then generated a score based on the phoneme transcriptions produced by each of our six model configurations.

The results, measured in Mean Absolute Error (MAE) and Mean Squared Error (MSE) against the ground-truth expert scores, are presented in Table~\ref{tab:llm-scoring-results}. The data reveals a clear and consistent trend: higher transcription quality directly leads to a more accurate automated score from the LLM, demonstrating a stronger agreement with human evaluators.

Specifically, the transcriptions from the fine-tuned ``Kids-FT'' models consistently result in lower MAE and MSE than those from the ``Base'' models. The optimal performance is achieved when the LLM is provided with the transcription from our best-performing model: the \textbf{Kids-FT} model paired with the flexible \textbf{WFST (K=2)} decoder. This configuration yielded the lowest error rates, with an \textbf{MAE of 8.43\%} and an \textbf{MSE of 0.2224}. This confirms that the detailed and accurate phoneme-level information captured by our proposed K-WFST framework provides the most robust and practically useful representation for downstream assessment tasks.

\begin{table}[h]
\centering
\caption{Performance of LLM-Assisted Scoring on the Multitudes ORF corpus.  Mean Absolute Error (MAE) and Mean
Squared Error (MSE) are calculated between the LLM-predicted scores and the manual expert scores. Lower values indicate higher agreement. The best results are highlighted in bold.}
\label{tab:llm-scoring-results}
\vspace{-2mm}
\begin{tabular}{llcc}
\toprule
\textbf{Model} & \textbf{Method} & \textbf{MAE (\%)} & \textbf{MSE} \\
\midrule
\multirow{3}{*}{\textbf{Base}}
& Greedy & 14.82 & 0.2876 \\
& WFST (K=1) & 11.78 & 0.2662 \\
& WFST (K=2) & 8.71 & 0.2371 \\
\midrule
\multirow{3}{*}{\textbf{Kids-FT}}
& Greedy & 10.29 & 0.2504 \\
& WFST (K=1) & 11.47 & 0.2581 \\
& WFST (K=2) & \textbf{8.43} & \textbf{0.2224} \\
\bottomrule
\end{tabular}
\vspace{-5mm}
\end{table}

%% file: text/conclusion.tex
K-Function unifies robust, child-oriented phoneme recognition via our K-WFST framework with sophisticated LLM reasoning into an end-to-end pipeline. This system converts children's speech into objective scores and valuable insights for language assessment. Its modular architecture shows strong promise for broad deployment across educational and interventional settings. Notably, K-WFST is inherently language-agnostic, as the SimMatrix relies on universal articulatory features adaptable to any language with defined phonetics. Future work will focus on expanding the framework to multilingual contexts, refining the analysis to finer linguistic units such as syllables, and verifying its long-term impact and fairness through large-scale field studies.

\section*{Acknowledgements}
Thanks for support from UC Noyce Initiative, Society of Hellman Fellows, NIH/NIDCD, and the Schwab Innovation fund.